%% file: 0.main.tex
\pdfoutput=1

\documentclass[11pt]{article}

\usepackage[final]{acl} 

\usepackage{times}
\usepackage{latexsym}

\usepackage[T1]{fontenc}

\usepackage[utf8]{inputenc}

\usepackage{microtype}

\usepackage{inconsolata}

\usepackage{graphicx}

\usepackage{booktabs}
\usepackage{multirow}
\usepackage{amsmath}
\usepackage{amssymb}
\usepackage{colortbl}   
\usepackage{makecell}
\usepackage{float}
\usepackage{placeins} 

\usepackage{graphicx}
\usepackage{subcaption}
\usepackage{lipsum}
\usepackage{pifont}
\usepackage[most]{tcolorbox}
\newtcolorbox{highlight}[1][]{
    enhanced,
    colback=yellow!10,
    colframe=gray!30,
    boxrule=0.5pt,
    arc=2pt,
    leftrule=0.5pt,
    rightrule=0.5pt,
    toprule=0.5pt,
    bottomrule=0.5pt,
    breakable,
    #1
}

\usepackage[normalem]{ulem}
\useunder{\uline}{\ul}{}

\title{Revisiting LLMs as Zero-Shot Time-Series Forecasters: \\ Small Noise Can Break Large Models  }

\author{
 \textbf{Junwoo Park$^{1,2}$},
 \textbf{Hyuck Lee$^{2}$},
 \textbf{Dohyun Lee$^{1,2}$},\\ 
 \textbf{Daehoon Gwak$^{1}$},
 \textbf{Jaegul Choo$^{1}$},\\
 \textsuperscript{1}KAIST AI \quad
 \textsuperscript{2}KRAFTON \quad
\\
 \texttt{\{junwoo.park,aiclaudev,daehoon.gwak,jchoo\}@kaist.ac.kr}\\
 \texttt{dlgur0921@krafton.com}\\
}

\begin{document}
\maketitle
\begin{abstract}
Large Language Models (LLMs) have shown remarkable performance across diverse tasks without domain-specific training, fueling interest in their potential for time-series forecasting. While LLMs have shown potential in zero-shot forecasting through prompting alone, recent studies suggest that LLMs lack inherent effectiveness in forecasting. Given these conflicting findings, a rigorous validation is essential for drawing reliable conclusions. In this paper, we evaluate the effectiveness of LLMs as zero-shot forecasters compared to state-of-the-art domain-specific models. Our experiments show that LLM-based zero-shot forecasters often struggle to achieve high accuracy due to their sensitivity to noise, underperforming even simple domain-specific models. 
We have explored solutions to reduce LLMs' sensitivity to noise in the zero-shot setting, but improving their robustness remains a significant challenge.
Our findings suggest that rather than emphasizing zero-shot forecasting, a more promising direction would be to focus on fine-tuning LLMs to better process numerical sequences. Our experimental code is available at \url{https://github.com/junwoopark92/revisiting-LLMs-zeroshot-forecaster}
\end{abstract}

\section{Introduction}
Large Language Models (LLMs) have demonstrated remarkable capabilities in diverse tasks, even without domain-specific training~\cite{kojima2022large,wei2022chain}. Building on this success, researchers are increasingly leveraging pre-trained LLMs for time series forecasting~\cite{zhou2023one,jin2023time,liu2024calf,gruver2024large, requeima2024llm,tang2025time}. A common approach involves converting the time series data into a format that LLMs can process, such as through alignment-tuning or by encoding sequences directly within prompts. This enables LLMs to perform few-shot or zero-shot forecasting. These studies show that LLM-based forecasters consistently perform well across multiple domains while reducing the need for domain-specific training. Notably, LLMTime~\cite{gruver2024large} demonstrated that LLMs can achieve zero-shot time series forecasting, performing at a level comparable to or even surpassing domain-specific models trained on the full dataset.

However, there remains an ongoing discussion about whether a thorough evaluation was conducted when integrating LLMs into time-series forecasting and assessing their performance. For instance, existing studies have used domain-specific models~\cite{zhou2021informer,zhou2022fedformer,wu2021autoformer} as baselines for comparison with LLMs, but these baseline models may not be sufficiently powerful~\cite{zeng2022transformers} to provide a fair assessment of LLM performance. Furthermore, a recent comprehensive study~\cite{tan2024language} found that LLMs do not significantly contribute to forecasting performance, undermining the justification for their high computational cost. These concerns discourage the adoption of LLMs for promising applications in time-series forecasting. Therefore, it is crucial to reassess the true performance of LLM-based forecasters. To address this, we raise the following question:

\begin{center} 
\fbox{ 
  \begin{minipage}{0.43\textwidth} 
  \centerline
   {\it Are LLMs as zero-shot forecasters truly}
   {\it more effective than domain-specific models?}
  \end{minipage}
}
\end{center}
Typically, zero-shot forecasting is useful because it does not require domain-specific training. However, if inference with a zero-shot model takes too much time, the end-to-end process of training and deploying a domain-specific model could be faster, which would undermine the main advantage of using a zero-shot approach. Therefore, to be considered a viable alternative, a zero-shot forecaster must at least offer faster inference than the total time required to train and deploy a domain-specific model. While inference speed is essential, accuracy remains equally important in scenarios where real-time prediction is not required. Domain-specific models often rely on high-quality training data, and their performance can be degraded in new domains with limited data. In such cases, pre-trained LLMs may offer better generalization. Therefore, even if an LLM is slower, a higher forecast accuracy can justify its use, making precision a valid criterion for assessing viability. 

In summary, for LLMs to be viable zero-shot forecasters, they should meet at least one of two key criteria. \textit{(1) they should generate predictions faster than domain-specific models that need to be trained for each new domain}, and \textit{(2) their forecasting accuracy should be superior to, if not at least comparable to, domain-specific models}. Even if only the first criterion is satisfied, meaning the model generates predictions faster than domain-specific models including their training time, it is still suitable for applications where real-time inference is important. Even if only the second criterion is satisfied, meaning the model achieves forecasting accuracy comparable to or better than domain-specific models, it remains useful in cases where performance is critical but immediate inference is not required.

In this paper, we thoroughly evaluate the effectiveness of LLM-based zero-shot forecasters by comparing their inference speed and accuracy against state-of-the-art~(SoTA) domain-specific forecasting models. Our experimental results indicate that LLM forecasters struggle to match the forecasting accuracy of domain-specific models. Furthermore, linear models, which can be trained on single input sequence, not only outperformed LLM-based forecasters in forecasting accuracy but also are more cost-efficient.
Additionally, our analysis reveals that the low accuracy of LLMs in time-series forecasting is closely tied to the presence of noise commonly found in real-world datasets. Despite exploring various strategies to mitigate this issue in zero-shot settings, our experiments show only marginal improvements, highlighting noise sensitivity as a fundamental limitation of LLM-based forecasters.

\vspace{-3mm}
\section{Experimental Setup}
\vspace{-2mm}
\input{Figures/exp_comparison_with_sota}
\noindent \textbf{Problem Definition\enskip}
Time series forecasting is the task of predicting future values based solely on past observations. Formally, given a uniformly sampled \( n \) observations of time series $s$ in which each observation has \(d\) dimensions, $s_n(t) = \{x_{t}, x_{t+1}, \dots, x_{t+n-1}\}, \quad x \in \mathbb{R}^{d}$, 
a forecaster predicts the future sequence $s_{O}(t+I)$ based on $s_{I}(t)$  where $I$ and $O$ denote an input length and output length of sequence, respectively. We define the application of a forecaster across multiple domains as follows: Given the \(N\) datasets \(\{D^1, \dots, D^N\}\), a forecaster predicts futures based on the most recent input sequence $s^i_I$ of each $D^i$. To solve this problem, we introduce three approaches: prompt-based zero-shot forecasting, domain-specific forecasting, and single-shot forecasting.

\noindent \textbf{Prompt-based LLM Forecasting Methods\enskip} Prompt-based zero-shot forecasting involves converting $s_I$ into a text sequence within a prompt and then asking LLMs to continue the sequence up to the desired length $O$. Existing methods, such as LLMTime and LLMP~\cite{requeima2024llm}, provide a foundation for LLM-based forecasting, but exploring additional prompting strategies can offer deeper insights into LLM capabilities. To this end, we designed TS-CoT and TS-InContext prompts (Figure~\ref{app:cot-incontext}), inspired by Chain of Thought (CoT) and In-Context Learning. We use both black-box (GPT-3.5, GPT-4, GPT-4o) and white-box models (LLaMA 2-7B, 3-8B, 2-70B, 3-70B, 3.1-70B).

\noindent \textbf{Domain-Specific Forecasting Models\enskip} \citet{gruver2024large} demonstrated that LLMTime outperforms conventional forecasting methods trained on the entire dataset, despite relying solely on the input sequence for prediction. However, most baseline models~\cite{wu2021autoformer,zhou2022fedformer} are regarded as ineffective baselines \cite{zeng2022transformers} and are relatively outdated compared to recently proposed forecasting models.
Therefore, to assess whether utilizing LLMs for forecasting is genuinely meaningful, we conduct comparisons with recent SoTA forecasting models~\cite{wu2022timesnet, nie2022time, liu2023itransformer, wang2024timemixer}. The characteristics of each model are detailed in Appendix~\ref{app:rel-domain}.

\noindent \textbf{Domain-Specific Single-shot Linear Models\enskip} Additionally, the linear models~\cite{zeng2022transformers} have been shown to achieve strong performance on widely used benchmarks despite having a small number of parameters. We found that these models can be trained on the same amount of data as LLMTime receives as input. Thus, we train linear models with only one input sequence and include them as single-shot linear models (details in Appendix~\ref{app:linear-detail}).

\noindent \textbf{Evaluation Protocol \enskip} To evaluate computational cost, we define the total computation cost $C$ of each approach as the sum of training cost $C^{\text{T}}$ and inference cost $C^{\text{I}}$. We assume that domain-specific models require separate training and inference for each dataset $D^i$, leading to $C_{\text{Domain}} = \frac{1}{N}\sum_{i=1}^{N}
  \Bigl[
    C^{\text{T}}\bigl(D^i\bigr)
    + C^{\text{I}}\bigl(s^i_I\bigr)
  \Bigr].
$
For pre-trained LLMs, we assume no additional training cost, so only the inference cost accumulates: $
C_{\text{LLM}}
= \frac{1}{N}\sum_{i=1}^{N}
  C^{\text{I}}\bigl(s^i_I\bigr).
$
Finally, the single-shot linear model also trained with a input sequence $s^i$ in each dataset independently before inference, resulting in
$
C_{\text{Linear}}
= \frac{1}{N}\sum_{i=1}^{N}
  \Bigl[
    C^{\text{T}}\bigl(\{s^i_I\}\bigr)
    + C^{\text{I}}\bigl(s^i_I\bigr)
  \Bigr].
$ We consider LLM forecasters cost-efficient as zero-shot forecasters only if $ C_{\text{LLM}} < C_{\text{domain}}$ and $ C_{\text{LLM}} < C_{\text{Linear}} $. Appendix~\ref{app:infertime} shows details for calculation of the inference time.
We measured the mean absolute error (MAE) and the mean squared error (MSE) to evaluate the forecasts.

\noindent \textbf{Benchmark Datasets \enskip} 
We evaluate all forecasters using three benchmark datasets: Monash~\cite{godahewa2021monash}, and Function~\cite{gruver2024large}, Informer~\cite{zhou2021informer} datasets. The Monash dataset consists of real-world time series from various domains. The Function dataset contains numerical sequences from mathematical functions to test pattern learning and extrapolation. The Informer dataset includes multivariate time-series data for long-sequence forecasting, having larger test set than others. Thus, to reduce LLM costs, we evaluate performance using only the last test sample. Appendix~\ref{app:dataset} provide details of the datasets.
\vspace{-2mm}
\section{Comparison with SoTA Domain-Specific Forecasters}
\vspace{-2mm}
\begin{highlight}
\paragraph{Finding 1:} 
\emph{SoTA domain-specific models are more accurate and cost-efficient than prompt-based LLM forecasters.}
\end{highlight}
We compare LLM-based forecasting models with SoTA domain-specific models (TimeMixer and iTransformer) in terms of computational efficiency and accuracy. As shown in Figure~\ref{fig:comp-sota}, LLMTime with LLaMA-3.1-70B performs comparably to forecasting models such as Autoformer and FEDformer, as evaluated in their original paper. However, it falls short compared to more recent SoTA models. Notably, LLM forecasters underperform compared to single-shot linear models~(DLinear-S and RLinear-S), despite these linear models being trained on just a single input sequence. Moreover, these linear models achieve significantly shorter inference time, even including training time, compared to LLMTime with LLaMA-2-7B. 

Due to space constraints, we provide additional experimental results in Appendix~\ref{app-sec:model-analysis}. Appendix~\ref{app-sec:hyper} presents an analysis of how different decoding settings (\textit{e.g.}, temperature and top-$p$ values) affect forecasting performance across various configurations, as summarized in Tables~\ref{tab:temp} and~\ref{tab:topp}. Appendix~\ref{app-sec:entire} extends the evaluation beyond the last-sample setting to cover the entire temporal horizon in univariate forecasting, as shown in Table~\ref{tab:entire_horizon}. Finally, Appendix~\ref{app-sec:more_baselines} provides additional baseline comparisons, including results for ARIMA~\cite{box1968some} and N-BEATS~\cite{oreshkin2019n}, alongside the linear models.

 
\subsection{Origin of LLMs Performance Degradation}
\begin{highlight}
\paragraph{Finding 2:} 
\emph{LLMs are highly sensitive to noise, while linear models remain robust.}
\end{highlight}

Another point of inquiry stems from the fact that LLMs have clearly demonstrated near-perfect performance on various functional datasets and accurate prediction results in the Monash dataset~\cite{gruver2024large}. In reproducing these experiments, we discovered that these datasets, unlike Informer datasets, contain very little noise and are clean. 

\input{Figures/exp_noise_function_dataset}
\noindent \textbf{Qualitative Results on the Function Dataset\enskip} For the function dataset, Figure ~\ref{fig:noise-func} shows how the performance of LLMs changes when a Gaussian noise is added to the original data. Even a very small amount of noise, relative to the original scale, can lead to significantly larger errors than the original predictions. This finding suggests that LLMs are highly sensitive to noise.

\input{Tables/exp_noise_injection_monash}
\noindent \textbf{Quantitative Results on the Monash Dataset\enskip} We also compare LLMs with single-shot linear models using the Monash dataset across various noise types, which are typically addressed in the noise robustness literature. Table~\ref{tab:noise-monash} shows that LLM-based methods significantly decline in performance compared to their original performance when exposed to three types of noise (details in Appendix~\ref{app:noise-type}). On the other hand, the linear models maintain nearly the same level of performance even under various types of noise. This result aligns with the theoretical findings of \citet{cheng2024robusttsf}, which suggest that models optimized with L1 and L2 loss exhibit robustness to noise. In addition to the three representative noise types discussed here, further evaluations under additional noise conditions—such as frequency-based noise—are presented in Appendix~\ref{app-sec:freq-noise}.

\subsection{Enhancing Noise Robustness of LLMs}
\begin{highlight}
\paragraph{Finding 3:} 
\emph{Additional input samples and noise filtering improve noise robustness of LLMs, but limited.}
\end{highlight}
\noindent \textbf{Increasing Input Length to Improve Noise Robustness \enskip}
\input{Figures/exp_increasing_input_length}
Because the robustness of conventional time-series models to noise depends on the number of training samples, we investigated whether providing longer input sequences could enhance the resilience of LLMs to noise. We hypothesized that using short input sequences might limit LLMs’ ability to distinguish noise; therefore, we evaluated their performance by gradually increasing the sequence length. As illustrated in Figure~\ref{fig:increasing-noise}, where the period in the Fourier series corresponds to an input sequence length, LLMs typically show only slight improvements in performance, whereas linear models gain significantly from longer sequences.


\input{Tables/exp_noise_filtering_informer}

\noindent \textbf{What If Applying Noise Filtering? \enskip}
The straightforward way to deal with noise is to remove it as a pre-processing for input. To observe how they affect performance, we applied two noise filtering techniques—Gaussian and Exponential Moving Average (EMA) filtering—on real-world datasets (details in Appendix~\ref{app:noisefilter}). As shown in Table~\ref{tab:noise-filter}, even after applying noise filtering techniques, LLM performance improves only slightly or remains unchanged.

\vspace{-1mm}
\section{Discussion}
\vspace{-1mm}
In this paper, we empirically demonstrate the limitations of LLM-based forecasting in real-world scenarios where noise is prevalent. A potential reason for LLMs' susceptibility to noise lies in their token-based encoding, which amplifies distortions in representation rather than reflecting actual numerical differences. This distortions hinder their ability to recognize meaningful patterns, making extrapolation more challenging. Given the strong reasoning capabilities of LLMs, we argue that rather than emphasizing zero-shot forecasting, a more promising research direction would be enhancing the capability to process numerical sequences through fine-tuning~\cite{hu2022lora, zha2022supervised, dettmers2023qlora}.

\section{Limitations}
Our study provides a comprehensive evaluation of LLM-based zero-shot forecasting, but there are areas that warrant further exploration. While we identify the sensitivity of LLMs to noise, the exact mechanisms behind this issue, such as the impact of tokenization and encoding, deserve deeper investigation. Our evaluation prioritizes computational cost and accuracy, yet aspects like interpretability, adaptability to diverse domains, and the potential benefits of integrating textual or multimodal data could offer valuable insights for future studies. Additionally, our experiments cover widely used benchmark datasets, but assessing LLM performance across an even broader range of real-world forecasting tasks would further strengthen our understanding of their capabilities and limitations.

\section*{Acknowledgments}
This work was supported by the Institute for Information \& Communications Technology Planning \& Evaluation (IITP) grant funded by the Korea government (MSIT) (RS-2019-II190075, Artificial Intelligence Graduate School Program at KAIST) and the National Research Foundation of Korea (NRF) grant funded by the Korea government (MSIT) (No. RS-2025-00555621).

\bibliography{custom}

\clearpage

\appendix

\section{Related Work}
\subsection{Domain-Specific Models for Time-series Forecasting}
\label{app:rel-domain}
Time-series forecasting has evolved from statistical methods to deep learning-based approaches. Informer~\cite{zhou2021informer} improved Transformer efficiency with ProbSparse self-attention, reducing computational overhead for long-term forecasting. Early models like LTSF-Linear~\cite{zeng2022transformers} demonstrated that simple linear regression could outperform complex architectures in structured datasets. PatchTST~\cite{nie2022time} introduced patching techniques inspired by vision models, enabling better locality retention. TimesNet~\cite{wu2022timesnet} leveraged temporal 2D-variation modeling, transforming time series into structured tensors for more efficient feature extraction. iTransformer~\cite{liu2023itransformer} addressed multivariate dependencies by applying attention mechanisms to variate tokens rather than time steps. TimeMixer~\cite{wang2024timemixer} introduced a fully MLP-based approach, effectively capturing multi-scale variations.

Compared to LLM-based methods, these domain-specific models optimize for numerical representation, long-term dependencies~\cite{park2024self}, and computational efficiency~\cite{xu2023fits}. While LLMs process time-series data as token sequences, they lack specialized multiscale decomposition and structured feature extraction found in these tailored models. Informer and PatchTST improve sparsity and efficiency, whereas iTransformer and TimesNet enhance multivariate correlation modeling. Linear models remain relevant, challenging the necessity of deep learning for structured time-series tasks. The progression from linear models to efficient deep learning architectures highlights the importance of designing task-specific solutions for forecasting. Future research should explore hybrid approaches that combine LLM flexibility with time-series-specific optimizations to further improve performance and efficiency.

\subsection{LLM-based Time-series Forecasting}
The use of LLMs for time series forecasting has recently gained attention, with two primary approaches emerging in the literature. One approach maps numerical sequences into the token embedding space of LLMs with minimal training, treating time series forecasting as a natural extension of next-token prediction. However, recent studies argue that this method does not effectively leverage the capabilities of LLMs for reasoning about time series. Notably, \citet{tan2024language} raise a crucial question: Are Language Models Actually Useful for Time Series Forecasting?. They challenge the assumption that LLM-based models outperform conventional forecasting methods, demonstrating that removing or replacing the LLM component often improves results. 

An alternative approach involves transforming numerical sequences into textual representations, allowing LLMs to interpret time series data through natural language prompts. This paradigm is exemplified by LLMTime~\cite{gruver2024large}, which was the first to introduce a zero-shot forecasting method using LLMs. Subsequently, LLMP~\cite{requeima2024llm} introduced modifications to prompt design, incorporating elements from stochastic processes to enhance predictive accuracy. 

Recent studies have increasingly emphasized the importance of textual information in forecasting tasks, demonstrating that integrating external knowledge sources can significantly enhance predictive performance. In particular, LLM-based approaches have shifted towards context-aided forecasting, where models utilize domain-specific textual inputs to inform future predictions. For example, \citet{gwak2024forecasting} show that LLMs can successfully predict future events from past news articles, underscoring their ability to capture temporal and contextual signals from language.

Building on this line of research, the CiK benchmark~\cite{williams2024context} provides a systematic framework for evaluating the role of auxiliary textual information in time-series forecasting. It demonstrates that models incorporating relevant domain knowledge significantly outperform purely numerical baselines. Moreover, CiK proposes evaluation protocols to assess how effectively LLMs integrate multimodal context, including both structured and unstructured inputs.
These developments suggest that future research should focus not just on zero-shot capabilities of LLMs, but on methods that enable them to synergistically leverage textual and numerical data for more robust forecasting.

\section{Experiment Details}

\subsection{Dataset Details}
\label{app:dataset}
\input{Tables/appendix_datasets}
To compare LLM forecasters with domain-specific models, we selected three commonly used datasets: Informer~\cite{zhou2021informer}, Monash~\cite{godahewa2021monash}, and Function datasets~\cite{gruver2024large}. The Informer dataset is a widely used benchmark for evaluating forecasting models, including multivariate time-series data from different domains such as ETTm2, Exchange Rate, Electricity, Traffic and Weather. This dataset mainly focuses on long-sequence forecasting. The Monash consists of public benchmark datasets from the Monash Time Series Forecasting Archive, covering various real-world domains such as energy consumption, traffic, and finance. Among these, we use eight time series from LLMTime~\cite{gruver2024large}. Function dataset contains numerical time series sampled from mathematical functions (\textit{e.g.,} sine, sigmoid, beat interference) to assess forecasting models’ ability to learn and extrapolate structured numerical patterns.

\subsection{Licenses and Terms of Use for Artifacts}
We utilize several datasets and pre-trained models, each with specific licensing terms. The Monash Time Series Forecasting (TSF) Repository is licensed under Creative Commons Attribution 4.0 International (CC BY 4.0), allowing research use with proper attribution\footnote{https://huggingface.co/datasets/Monash-University/monash\_tsf}
The Informer datasets do not explicitly state a license, but users are advised to check the official repository for terms of use\footnote{https://github.com/zhouhaoyi/Informer2020}. The Function dataset similarly lacks stated licensing details, requiring direct consultation with the authors. The study also employs pre-trained models, including GPT-3.5 and GPT-4, which are governed by OpenAI’s proprietary API terms, allowing usage under strict conditions\footnote{https://openai.com/terms}. LLaMA models, provided by Meta AI, are subject to a research-focused license, permitting non-commercial research use\footnote{https://ai.facebook.com/blog/large-language-model-llama-meta-ai}. Users must comply with each artifact’s licensing conditions, particularly regarding research-only restrictions and commercial use limitations.

\noindent \textbf{AI Assistants in Research or Writing \enskip}
We utilized AI assistants, specifically ChatGPT, to support research, writing, and coding. The AI was used for tasks such as drafting text, summarizing information, and refining explanations. This usage is documented to ensure transparency and acknowledgment of AI contributions.

\subsection{Details of Comparison with SoTA Domain-Specific Models}
\label{app:eval-proto}
\noindent \textbf{Evaluation Protocol \enskip}
A widely used evaluation method for time-series forecasting involves splitting the entire series into training, validation, and test sets while maintaining the chronological order. Specifically, the most recent 20\% of the time series is designated as the test set, while the remaining portion is used for training and hyperparameter tuning through validation. Given that time-series forecasting inherently involves predicting future values, it is crucial to split the dataset sequentially to prevent data leakage.

Once the dataset is split into these temporal segments, a sliding window approach is employed, where a fixed-length window moves with a stride through the time series. Each window consists of an input sequence of length \(I\) and an output sequence of length \(O\), which serves as the ground truth for evaluating model predictions. The model is trained by minimizing the loss between its predicted output sequence and the actual values. Table 3 provides statistical information about the datasets used, including their total length and the number of samples obtained after applying the sliding window method.

\noindent \textbf{Last Sample Evaluation \enskip}
While this evaluation method is suitable for domain-specific models, it introduces significant computational and financial costs when applied to LLM-based approaches due to the increased number of API calls and processing requirements. To mitigate this burden, LLMTime employs a reduced evaluation strategy, selecting only the last sample from the Informer dataset for testing. Consequently, the number of test samples in each domain is proportional to the number of channels in that domain. Since LLMTime does not disclose the exact methodology used for this selection, our evaluation follows a similar approach by selecting the last available sample from the conventional time-series split. While this approach prevents direct comparison with results from the original paper, we confirm that the general trend of the findings remains consistent.

\noindent \textbf{Domain-Specific Model Details \enskip}
In the field of time-series forecasting, numerous models have emerged since the introduction of Informer, each showcasing strong performance with distinct innovations. We concentrate on evaluating the most advanced and high-performing models from recent developments. Our approach is validated against recent seven forecasting baselines, with all models implemented in PyTorch. For the latest forecasting models—including DLinear\footnote{https://github.com/cure-lab/LTSF-Linear}, MICN\footnote{https://github.com/wanghq21/MICN}~\cite{wang2023micn}, and PatchTST\footnote{https://github.com/yuqinie98/PatchTST}~\cite{nie2022time}, TimesNet~\cite{wu2022timesnet}, iTransformer~\cite{liu2023itransformer}, TimeMixer\footnote{https://github.com/thuml/Time-Series-Library}~\cite{wang2024timemixer}—we utilized the official implementations provided by the original authors rather than developing them independently.
For domain-specific models, we conducted five independent training and inference runs, varying the random seed in each run. 

\noindent \textbf{LLM-based Model Details\enskip}
For LLM-based models, we performed five inference runs on the same test sample and computed the median of the five predictions to mitigate the impact of stochastic variations due to temperature settings. This median aggregation method is commonly used in LLM-based models to reduce the affect of outlier predictions. As a result, domain-specific models report confidence intervals in the results, while LLM-based models do not due to the fivefold increase in computational cost that would be required. The API model utilizes the API provided by OpenAI~\footnote{https://api.openai.com/v1/}, while all LLaMA models use pretrained models provided by Hugging Face~\footnote{https://huggingface.co/meta-llama}.

\noindent \textbf{Training Strategy for Linear-S Models \enskip}
\label{app:linear-detail}
\input{Tables/desc_linear_n_data}
Linear-S models follow a different training and inference process. These models use a single input sequence only for training. To conduct both training and validation with a single sample, we need to create a windowed dataset from the input sequence, similar to when training a domain-specific model. Given an input sequence length \( I \) and a target prediction length \( O \), we transform this into a smaller input length \( I' \) and output length \( O' \) to construct a windowed dataset. The total number of windows \( K \) is determined as follows:
\begin{equation}
K = d(I - (I' + O') + 1)
\end{equation} where $d$ is the number of channels. The linear model employs a channel independence strategy~\cite{nie2022time}, where channels are transformed into the batch axis, allowing for independent predictions across channels. As a result, in multivariate forecasting, the number of windows is proportional to the number of channels. In the sliding window framework, if \( K \) is too small, training and validation of the linear model become challenging. Therefore, with \( I \), \( O \), and \( d \) fixed, we need to adjust \( I' \) and \( O' \) to ensure a sufficient number of windowed data points. If \( O' \) is smaller than \( O \), inference must be performed in an autoregressive manner, which can lead to accumulated errors and degrade performance. Considering these characteristics, we set $I'$ and $O'$ as half of $I$ and $O'$, respectively. The statistics of each dataset are presented in Table~\ref{tab:linear-windows}.

\noindent \textbf{Total Inference Time Calculation \enskip}
\label{app:infertime}
To compare the inference speeds of domain-specific models and LLM-based models, we measure the total time required for a single prediction. For domain-specific models, this total time includes the training time (utilizing both the training and validation sets) and the inference time required to generate a single prediction. For API-based LLM models~(\textit{e.g.,} GPT-3.5, GPT-4 and GPT-4o), the total inference time includes the duration required to query the API five times and compute the median prediction. Since these queries are processed in asynchronous threading, The runtime for executing five queries is not five times the runtime of a single query. Moreover, reducing the number of queries to a single run results in significantly degraded performance, making this trade-off necessary. For Local-deployed LLM models~(\textit{e.g.,} LLaMA-2-70B, LLaMA-3.1-70B), we measured the time taken to perform a batch of five inference queries. Due to the 40GB GPU memory limitation, the 70B model was loaded in 4-bit mode for inference. All local models, except for the API model, were executed in the same computing environment with an A100-40GB GPU.

\subsection{Details of Noise Experiments}
\label{app:noise-type}
\noindent \textbf{Noise Types \enskip}
\citet{cheng2024robusttsf} categorize three primary types of noise in time series forecasting: Constant, Missing and Gaussian.
Constant noise introduce a fixed deviation from the ground-truth value. These noise can be mathematically represented as:
\begin{equation}
    z_A = z + \epsilon,
\end{equation}
where $z_A$ is the observed value with anomaly, $z$ is the true value, and $\epsilon$ is a constant perturbation.
Missing noise correspond to completely missing values in the dataset. These noise are defined as:
\begin{equation}
    z_A = \epsilon,
\end{equation}
where $\epsilon$ is a constant, often set to zero.
Gaussian noise represent deviations that follow a Gaussian distribution. These noise are defined as:
\begin{equation}
    z_A = z + \epsilon, \quad \epsilon \sim \mathcal{N}(0, \sigma^2),
\end{equation}
where $\epsilon$ is drawn from a normal distribution with mean $0$ and variance $\sigma^2$. We conducted experiments by injecting three types of noise into the Monash dataset to evaluate the robustness of the LLM to each type of noise.

The single-shot linear models exhibit greater robustness to noise compared to LLM-based models due to the statistical properties of their loss functions. The L1 loss function maintains a constant sum of losses under clean and noisy conditions, ensuring that optimization remains unaffected by noise. Theoretical results~\cite{cheng2024robusttsf} support our empirical findings. Since LLMs are token-based models and utilized as zero-shot forecasters, achieving noise robustness is more challenging for them.

Beyond the three representative types of noise discussed above, time-series data often involve more complex challenges such as changes in trend and variations in frequency~\cite{park23deep,zhou2024llmsunderstandtimeseries}. To address this, we additionally conducted experiments comparing LLM-based models and domain-specific models under scenarios involving frequency changes as shown in Table~\ref{tab:monash_freq_noise}.

\noindent \textbf{Noise Filtering Methods \enskip}
\label{app:noisefilter}
Gaussian filtering and EMA filtering are both widely used techniques for noise reduction in signal and image processing. Gaussian filtering applies a Gaussian kernel to smooth data by averaging neighboring values with weights determined by a normal distribution, effectively reducing high-frequency noise while preserving important structures. It is commonly used in image processing to blur and remove noise. On the other hand, EMA filtering is a recursive technique that assigns exponentially decreasing weights to past observations, allowing recent data points to have more influence. This makes EMA particularly effective for real-time applications, such as financial analysis and sensor data smoothing, where responsiveness to recent changes is crucial while still maintaining noise reduction. 

However, Gaussian filtering can overly smooth time series data, potentially removing important short-term variations and trends. EMA filtering, while responsive to recent changes, may lag behind true values and be sensitive to sudden spikes. Both methods struggle with non-stationary data, where patterns and noise characteristics change over time. Therefore, applying such noise filtering to a time series before inference may degrade the original dynamics and may not necessarily improve performance.

\vspace{-2mm}
\section{Additional Results}
\vspace{-2mm}
\subsection{Comparison Between Prompt Methods}
Based on the findings of \citet{wei2022chain} and \citet{kojima2022large}, Chain-of-Thought (CoT) prompting plays a crucial role in enhancing the reasoning capabilities of large language models. \citet{wei2022chain} demonstrated that explicitly structuring the reasoning process in prompts allows models to perform complex, multi-step reasoning more effectively. Meanwhile, \citet{kojima2022large} showed that even simple prompts like "Let's think step by step" can significantly improve zero-shot reasoning. Building on these insights, when designing a prompt for time-series forecasting, it is essential to incorporate CoT principles. This involves guiding the model through a structured, step-by-step approach to analyzing trends, identifying patterns, and making predictions based on historical data. By explicitly prompting the model to reason through the forecasting process, we design TS-CoT Prompt (Figure~\ref{app:cot-incontext}) to improve the accuracy and reliability of predictions in time-series forecasting.

Building on the insights from \citet{brown2020language}, In-Context Learning (ICL) serves as a powerful mechanism for improving downstream tasks when using LLMs~\cite{zelikman2022star,peng2024regenesis}. These studies introduced the concept of few-shot learning, demonstrating that language models can generalize from a limited number of examples without fine-tuning. \citet{zelikman2022star} extended this idea with STaR, enabling self-improvement through iterative reasoning, while \citet{peng2024regenesis} proposed ReGenesis, a framework for continual self-improvement via model-generated feedback. By leveraging these principles, we design TS-InContex prompt (Figure~\ref{app:cot-incontext}), which provides a structured set of in-output sequences within the prompt itself, allowing the model to recognize patterns. 
\label{app:comp-prompt}
\input{Tables/appendix_comp_prompts}
\input{Figures/desc_LLMTime_Prompt}
\input{Figures/desc_LLMP_Prompt}
\input{Figures/desc_CoT_InContext}

\vspace{-2mm}
\subsection{Model Analysis}
\label{app-sec:model-analysis}
\subsubsection{LLMs' Hyperparameter Analysis}
\label{app-sec:hyper}
\input{Tables/exp_temperature}
\input{Tables/exp_topp}

\paragraph{Temperature.} We adopt the temperature values reported by LLMTime, which were selected to optimize zero-shot forecasting performance. To assess the sensitivity of LLM performance to this hyperparameter, we conduct an ablation study varying the temperature during inference. Table~\ref{tab:temp} presents the MAE and MSE scores across three datasets.

On ETTm2 and Weather, the default temperature (1.0) yields the best performance. For Exchange Rate, a lower temperature (0.5) marginally improves the error metrics but still fails to outperform linear baselines. These results indicate that while temperature affects performance to a limited degree, the general trend persists: LLMs consistently underperform compared to domain-specific models across all datasets.

\paragraph{Top-p Sampling.}
Top-p (nucleus) sampling is a widely adopted decoding strategy in modern LLMs, striking a balance between coherence and diversity~\cite{holtzman2019curious,nguyen2024turning}. It is the default in models such as GPT-4o and LLaMA, and is also used in LLMTime. We perform an ablation study to evaluate how varying the Top-p parameter impacts zero-shot forecasting.

As shown in Table~\ref{tab:topp}, the default value (0.8) leads to strong performance among the tested configurations. However, consistent with the temperature study, GPT-4o remains inferior to linear models across all datasets, suggesting that decoding hyperparameters do not fundamentally alter the performance gap.

\subsubsection{Evaluating Entire Horizon on Three Real-world Datasets}
\label{app-sec:entire}
\input{Tables/exp_entire_horizon}
We adopt the last-sample evaluation protocol introduced by prior work (LLMTime), which evaluates forecasting models only on the final prediction window within each test sequence. This approach substantially reduces the computational and financial costs of repeated LLM inference, particularly when relying on commercial APIs. It is important to clarify that the term last sample does not refer to a single data point, but rather to the final forecasting window for each multivariate time series in the test split. Since the LLM-based forecasters operate independently on each time series, the total number of forecast instances evaluated under this setting is 1,219, as detailed in Appendix Table~\ref{tab:data-stat}.

To provide a more complete evaluation of temporal generalization, we further conduct experiments using a sliding-window evaluation protocol. In this setup, forecasts are generated across all time steps in the test split by sliding the input window with a stride equal to the prediction length. For consistency and interpretability, we focus our analysis on the main target variable of each dataset (\textit{e.g.,} the oil temperature variable in ETTm2).

As shown in Table~\ref{tab:entire_horizon}, the results from this extended evaluation confirm our earlier findings: while LLM-based forecasters such as GPT-4o and LLaMA3.1-8B-Instruct exhibit competitive short-term performance, linear models (\textit{e.g.,} DLinear-S, RLinear-S) continue to outperform them in terms of robustness and accuracy across the entire test horizon. These observations reinforce the limitations of current zero-shot LLM forecasters and underline the temporal stability offered by lightweight, domain-specific architectures.

\subsubsection{Evaluating Robustness Against Frequency-Based Noise}
\label{app-sec:freq-noise}
\input{Tables/exp_freq_noise}
To further investigate the robustness of forecasting models under structured perturbations, we conduct additional experiments using periodic noise. Two variants of noise are considered: (1) Freq-Add, in which a sinusoidal component is added to the original time series, and (2) Freq-Replace, where the original series is replaced entirely by a sinusoidal signal. These settings are designed to emulate scenarios where time-series data are affected by periodic interference unrelated to the true signal.

Experimental results in Table~\ref{tab:monash_freq_noise} show that LLM-based models, such as GPT-4 and GPT-4o, experience substantial performance degradation in the presence of such periodic distortions. In contrast, linear baselines such as DLinear-S and RLinear-S remain largely unaffected, demonstrating strong resilience to this form of noise. This observation aligns with previous findings, reinforcing that linear models exhibit superior robustness when exposed to structured, non-informative patterns in the input.

\subsubsection{Comparison with Additional Baselines}
\label{app-sec:more_baselines}
\input{Tables/exp_more_baselines}
Statistical and lightweight forecasting models, such as ARIMA~\cite{box1968some} and N-BEATS~\cite{oreshkin2019n}, are widely recognized for their simplicity and computational efficiency. To ensure a comprehensive evaluation, we included these models in our benchmark and report their forecasting performance on three real-world datasets in Table~\ref{tab:more_baselines}. The results show that, while LLM-based zero-shot models (\textit{e.g.,} GPT-4o) can outperform ARIMA and N-BEATS in certain settings, their performance is still surpassed by simpler linear models such as DLinear-S and RLinear-S trained on a single input sequence. These single-shot linear models not only achieve higher forecasting accuracy but also retain strong computational efficiency. We propose these models as new, competitive baselines for zero-shot forecasting with LLMs, providing a more challenging yet realistic reference point for future research.


\clearpage
\onecolumn
\subsection{Qualitative Results of Forecasting Models}
\input{Figures/appendix_Electricity_vis}
\input{Figures/appendix_ETTm2_vis}
\input{Figures/appendix_ExchangeRate_vis}
\input{Figures/appendix_Weather_vis}
\input{Figures/appendix_Traffic_vis}
\clearpage
\onecolumn
\subsection{Full Results}
\input{Figures/appendix_full_results}

\end{document}

%% file: Figures/exp_comparison_with_sota.tex
\begin{figure*}[h!]
\begin{center}
  \includegraphics[width=1.0\linewidth]{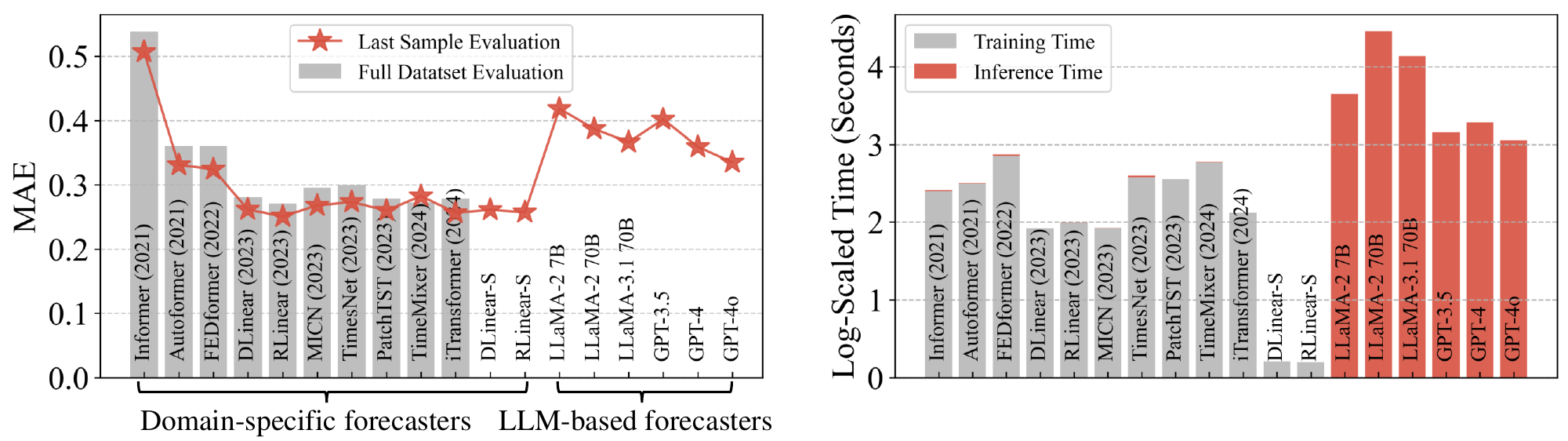}
\end{center}
\vspace{-3mm}
\caption{Multivariate forecasting results on five datasets from the Informer benchmark ($I$=$\{96, 384\}$ and $O$=$\{48,192\}$). We report the averaged MAE and log-scaled inference time for domain-specific forecasters and LLM-based forecasters using the best-performing LLMTime prompt (results for other three prompts are provided in Table~\ref{apptab:comp-prompts}). (Left) Recent domain-specific forecasters achieve lower MAE than LLM-based forecasters in last sample evaluation~\cite{gruver2024large}. (Right) Overall, LLM-based forecasters have longer inference times than domain-specific models, despite the latter requiring domain-specific training before inference. Moreover, single-shot linear models (DLinear-S and RLinear-S), trained solely on the input sequence of each domain, achieve significantly shorter inference times and outperform LLM-based forecasters. Appendix~\ref{app:eval-proto} includes a detailed evaluation protocol. Figure~\ref{fig:app-ecl-vis} shows qualitative results and Figure~\ref{appfig:full-res} includes full results with confidence intervals.}
\label{fig:comp-sota}
\vspace{-2mm}
\end{figure*}

%% file: Figures/exp_noise_function_dataset.tex
\begin{figure*}[h!]
\begin{center}
  \includegraphics[width=1.0\linewidth]{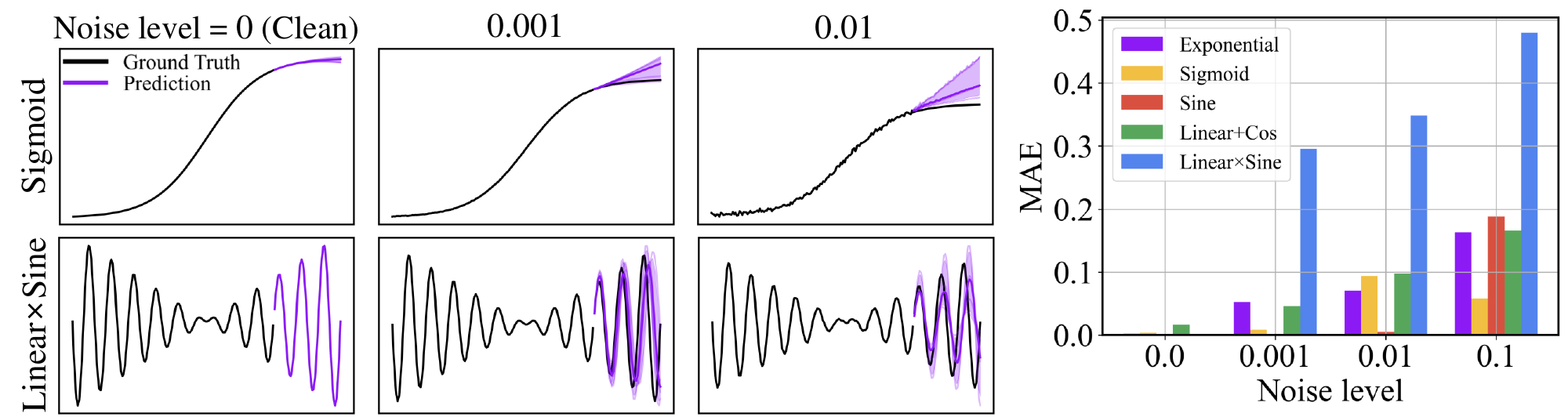}
\end{center}
\vspace{-3mm}
\caption{Performance variations in the Function dataset based on noise levels. LLMTime (GPT-4) performed perfectly on clean data, but even a slight noise (\textit{e.g.,} Gaussian noise with a standard deviation of 0.001) addition within the 0-1 input scale significantly increased MAE.}
\label{fig:noise-func}
\vspace{-3mm}
\end{figure*}

%% file: Tables/exp_noise_injection_monash.tex
\begin{table}[]
\resizebox{\columnwidth}{!}{
\begin{tabular}{@{}l|rr|rr|rr|rr@{}}
\toprule
\multicolumn{1}{c}{\multirow{2}{*}{\textbf{Models}}} & \multicolumn{2}{c}{\textbf{Clean}}                & \multicolumn{2}{c}{\textbf{Gaussian}}             & \multicolumn{2}{c}{\textbf{Constant}}             & \multicolumn{2}{c}{\textbf{Missing}}              \\ \cmidrule(l){2-9} 
\multicolumn{1}{c}{}                       & \multicolumn{1}{c}{MAE} & \multicolumn{1}{c|}{MSE} & \multicolumn{1}{c}{MAE} & \multicolumn{1}{c|}{MSE} & \multicolumn{1}{c}{MAE} & \multicolumn{1}{c|}{MSE} & \multicolumn{1}{c}{MAE} & \multicolumn{1}{c}{MSE} \\ \midrule
GPT-3.5                                    & 0.8247                  & 1.4464                  & 0.9215                  & 1.6484                  & 0.9018                  & 1.4923                  & 0.9897                  & 1.7561                  \\
LLAMA2 7B                                   & 0.8875                  & 1.5366                  & 0.8831                  & 1.4087                  & 0.9380                  & 1.6075                  & 0.9482                  & 1.4286                  \\
LLAMA3 8B                                   & 0.7205                  & 1.0185                  & 1.0307                  & 1.9737                  & 0.8974                  & 1.4575                  & 1.0004                  & 1.5850                  \\
GPT-4                                      & \textbf{0.6956}         & 0.9533                  & 0.7330                  & 1.0234                  & 0.7851                  & 1.1315                  & 0.9115                  & 1.4298                  \\
LLAMA2 70B                                  & 0.7196                  & 1.0028                  & 0.7776                  & 1.2476                  & 0.8773                  & 1.3353                  & 0.8404                  & 1.2749                  \\
LLAMA3 70B                                  & 0.7006                  & 0.9596                  & 0.8870                  & 1.5695                  & 0.8184                  & 1.1474                  & 0.8461                  & 1.1829                  \\ \midrule
DLinear-S                                  & {\ul 0.7074}            & \textbf{0.8890}         & \textbf{0.7063}         & \textbf{0.8896}         & \textbf{0.7120}         & \textbf{0.8997}         & \textbf{0.7112}         & \textbf{0.8910}         \\
RLinear-S                                  & 0.7182                  & {\ul 0.8804}            & {\ul 0.7196}            & {\ul 0.8817}            & {\ul 0.7171}            & {\ul 0.8766}            & {\ul 0.7438}            & {\ul 0.8971}            \\ \bottomrule
\end{tabular}
}
\caption{When the Monash dataset contains Gaussian, Constant, and Missing noise, commonly used to evaluate noise robustness~\cite{cheng2024robusttsf}, LLM-based forecasters experience a performance decline, while Linear-S models perform similarly to clean data.}
\label{tab:noise-monash}
\vspace{-6mm}
\end{table}

%% file: Figures/exp_increasing_input_length.tex
\begin{figure}[h!]
\begin{center}
  \includegraphics[width=1.0\linewidth]{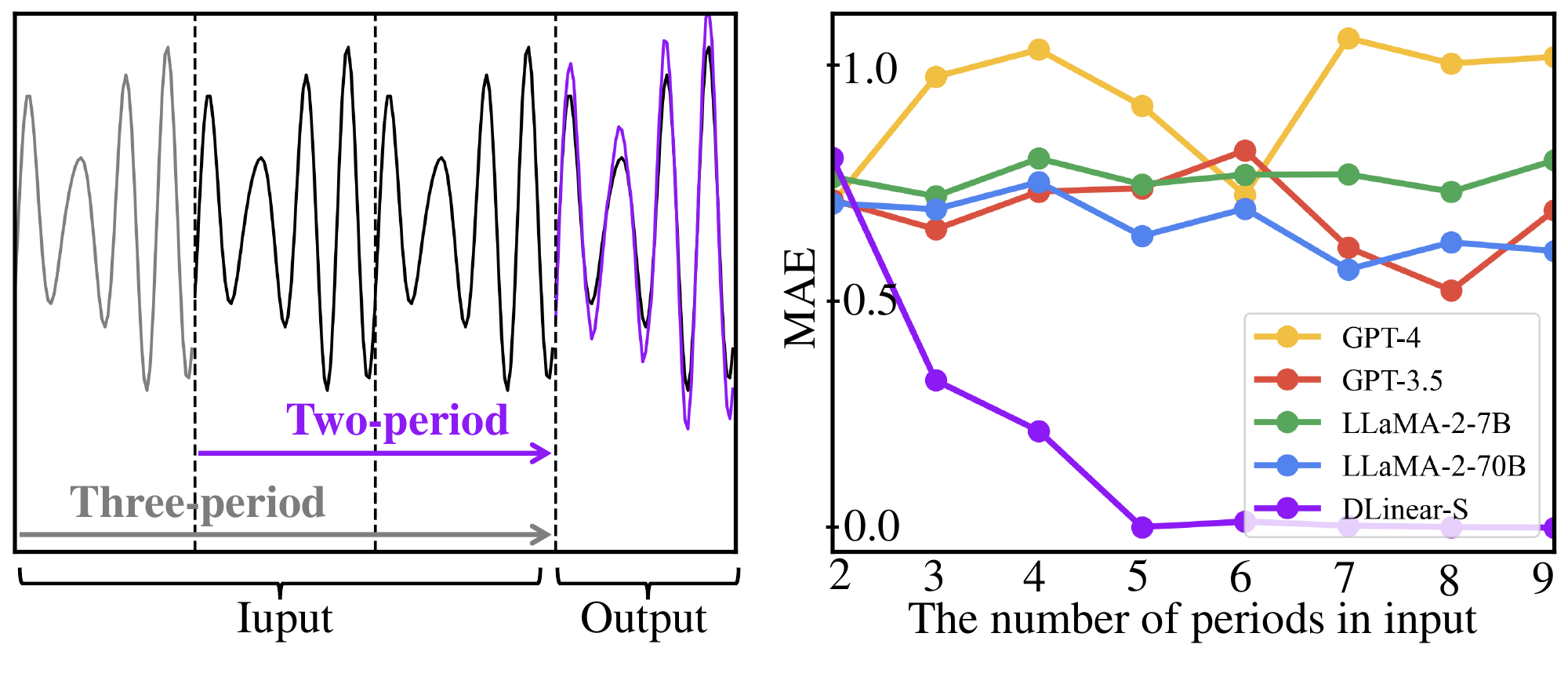}
\end{center}
\vspace{-5mm}
\caption{Increasing the input sequence length in prompts for LLM-based models leads to a slight performance improvement, but the gains remain minimal compared to DLinear-S.}
\label{fig:increasing-noise}
\vspace{-3mm}
\end{figure}

%% file: Tables/exp_noise_filtering_informer.tex
\begin{table}[]
\resizebox{\columnwidth }{!}{
\begin{tabular}{@{}l|l|rr|rr|rr@{}}
\toprule
\multicolumn{1}{c}{\multirow{2}{*}{\textbf{Models}}} & \multicolumn{1}{c}{\multirow{2}{*}{\textbf{Filtering}}} & \multicolumn{2}{c}{\textbf{ETTm2}}                  & \multicolumn{2}{c}{\textbf{Exchange Rate}}                   & \multicolumn{2}{c}{\textbf{Weather}}                \\ \cmidrule(l){3-8} 
\multicolumn{1}{c}{}                                            & \multicolumn{1}{c}{}                                    & \multicolumn{1}{c}{MAE} & \multicolumn{1}{c|}{MSE} & \multicolumn{1}{c}{MAE} & \multicolumn{1}{c|}{MSE} & \multicolumn{1}{c}{MAE} & \multicolumn{1}{c}{MSE} \\ \midrule
GPT-4o                                                 & X                                                       & 0.2322                   & 0.1237                   & 0.3595                   & 0.2893                            & 0.1880                   & 0.1147                   \\
GPT-4o                                                 & Gaussian                                                &  \cellcolor{pink} {\ul 0.1995}          & \cellcolor{pink} 0.1097          & 0.3718                   &  \cellcolor{pink} 0.2643                   &  \cellcolor{pink} 0.1698          &  \cellcolor{pink} 0.0827          \\
GPT-4o                                                 & EMA                                                     & 0.2501                   & 0.1519                   & 0.3615                   &  \cellcolor{pink} 0.2882                   &   \cellcolor{pink} 0.1808                   &   \cellcolor{pink} 0.0972          \\
\midrule
LLAMA3.1 70B                                           & X                                                       & 0.2462                   & 0.1579                   & 0.4646                   & 0.3596                            & 0.1659                   & 0.0741                   \\
LLAMA3.1 70B                                          & Gaussian                                                &0.2659     & 0.1868     &  \cellcolor{pink} 0.4323     &   \cellcolor{pink} 0.3579              & 0.1735     & 0.0981     \\
LLAMA3.1 70B                                          & EMA                                                     &  \cellcolor{pink} 0.2431     & 0.1661     &  \cellcolor{pink} 0.3765     &  \cellcolor{pink} 0.2782              & 0.1689     & 0.0955     \\
\midrule
DLinear-S                                            & X                                                       & 0.2039                   & {\ul 0.0782}                   & \textbf{0.2296}          & \textbf{0.1166}                   & {\ul 0.1643}                   & {\ul 0.0522}                   \\
RLinear-S                                             & X                                                       & \textbf{0.1824}          & \textbf{0.0677}          & {\ul 0.2571}                   & {\ul 0.1370}                             & \textbf{0.1392}          & \textbf{0.0469}          \\ \bottomrule
\end{tabular}
}
\caption{When two noise filtering methods (Gaussian and EMA) are applied to LLM-based models on the Informer dataset, performance improves (highlighted in \colorbox[rgb]{1.0, 0.752, 0.796}{\parbox[c][2pt][c]{4mm}{red}}). Nevertheless, single-shot linear models consistently outperform LLM-based models.}
\label{tab:noise-filter}
\vspace{-6mm}
\end{table}

%% file: Tables/appendix_datasets.tex
\begin{table}[]
\resizebox{\columnwidth}{!}{
\begin{tabular}{l|l|l|l|l}
\toprule
\textbf{Dataset} & \textbf{Domain}    & \textbf{Length} & \textbf{Channels} & \textbf{frequency} \\ \midrule
ETTm2            & System Monitoring  & 69680                       & 7                               & 15 minutes                  \\ \midrule
Electricity      & Energy Consumption & 26304                       & 321                             & 1 Hour                      \\ \midrule
Traffic          & Traffic            & 17544                       & 862                             & 1 Hour                      \\ \midrule
Weather          & Weather            & 52695                       & 21                              & 10 Minutes                  \\ \midrule
ExchangeRate         & Economic           & 7588                        & 8                               & 1 Day                       \\ 
\midrule
Monash Dataset             &  Diverse domains             &      [144:900]            & 8                               & -         
\\
\midrule
Function Dataset              & Mathematical Equations              & 200                         & 6                               & -  
\\
\bottomrule
\end{tabular}
}
\caption{Statistics of Informer, Monash, and Function datasets. In the last sample evaluation, we compare the performance of each model by taking the last $I + O$ length from each dataset. Since LLMs predict each channel independently in multivariate forecasting, the number of sequences that LLMs need to predict is equal to the number of channels.}
\label{tab:data-stat}
\vspace{-3mm}
\end{table}

%% file: Tables/desc_linear_n_data.tex
\begin{table}[]
\resizebox{\columnwidth }{!}{
\begin{tabular}{@{}l|r|r|r|r|r|r@{}}
\toprule
\textbf{Informer Dataset}       & \multicolumn{1}{c|}{\textbf{$d$}} & \multicolumn{1}{c|}{\textbf{$I$}} & \multicolumn{1}{c|}{\textbf{$O$}} & \multicolumn{1}{c|}{\textbf{$K$}} & \multicolumn{1}{c|}{\textbf{$I'$}} & \multicolumn{1}{c}{\textbf{$O'$}} \\ \midrule
ETTm2         & 7                      & 384                    & 192                    & 1,351                  & 96                      & 96                      \\ \midrule
Exchange Rate & 8                      & 384                    & 192                    & 1,544                  & 96                      & 96                      \\ \midrule
Weather       & 21                     & 384                    & 192                    & 4,053                  & 96                      & 96                      \\ \midrule
Electricity   & 321                    & 96                     & 48                     & 15,729                 & 24                      & 24                      \\ \midrule
Traffic       & 862                    & 96                     & 48                     & 42,238                 & 24                      & 24                      \\ \bottomrule
\end{tabular}
}
\caption{The number of windows $K$ for training and validating the single-shot linear models according to the internal prediction length $O'$ and input length $I'$ for each dataset with the target input $I$ and output $O$ lengths.}
\label{tab:linear-windows}
\vspace{-3mm}
\end{table}

%% file: Tables/appendix_comp_prompts.tex
\begin{table}[]
\resizebox{\columnwidth }{!}{
\begin{tabular}{@{}l|c|rr|rr|rr@{}}
\toprule
\multicolumn{1}{c}{\multirow{2}{*}{\textbf{Prompt}}} & \multicolumn{1}{c}{\multirow{2}{*}{\textbf{Input Length}}} & \multicolumn{2}{c|}{\textbf{ETTm2}}                & \multicolumn{2}{c|}{\textbf{Exchange Rate}}        & \multicolumn{2}{c}{\textbf{Weather}}              \\ \cmidrule(l){3-8} 
\multicolumn{1}{c}{}                        & \multicolumn{1}{c}{}                              & \multicolumn{1}{c}{MAE} & \multicolumn{1}{c|}{MSE} & \multicolumn{1}{c}{MAE} & \multicolumn{1}{c|}{MSE} & \multicolumn{1}{c}{MAE} & \multicolumn{1}{c}{MSE} \\ \midrule
\multirow{3}{*}{\shortstack{LLMTime \\ (b of Figure~\ref{app:llmtime-prompt})}}                     & 384                                                & \textbf{0.2322}                  & \textbf{0.1237}                   & 0.3595                  & 0.2893                   & \textbf{0.1880}                  & \textbf{0.1147}                  \\
                                             & 576                                                & 0.2741                  & 0.2037                   & 0.3323                  & 0.2539                   & 0.1943                  & 0.1231                  \\
                                             & 768                                                & 0.3298                  & 0.2539                   & 0.3807                  & 0.4030                   & 0.2075                  & 0.1398                  \\ \midrule
\multirow{3}{*}{\shortstack{LLMP \\ (b of Figure~\ref{app:LLMP-prompt})}}                        & 384                                                & 0.2860                  & 0.1969                   & 0.3436                  & 0.2869                   & 0.2039                  & 0.1200                  \\
                                             & 576                                                & 0.3060                  & 0.2188                   & \underline{0.3336}                  & \underline{0.2169}                   & \underline{0.1903}                  & \underline{0.1167}                  \\
                                             & 768                                                & \underline{0.2471}                  & \underline{0.1388}                   & 0.4171                  & 0.3855                   & 0.1914                  & 0.1197                  \\ \midrule
\multirow{3}{*}{\shortstack{TS-CoT \\ (a of Figure~\ref{app:cot-incontext})}}                      & 384                                                & 0.2823                  & 0.1799                   & 0.3647                  & 0.2358                   & 0.1950                  & 0.1451                  \\
                                             & 576                                                & 0.3732                  & 0.3355                   & 0.3297                  & 0.2227                   & 0.2222                  & 0.1310                  \\
                                             & 768                                                & 0.2941                  & 0.2075                   & 0.4083                  & 0.3284                   & 0.1831                  & 0.1666                  \\ \midrule
\multirow{3}{*}{\shortstack{TS-InContext \\ (b of Figure~\ref{app:cot-incontext})}}                & 384 (1-Shot)                                               & 0.2933                  & 0.1909                   & 0.3296                  & 0.2111                   & 0.1921                  & 0.1283                  \\
                                             & 576 (2-Shot)                                               & 0.2678                  & 0.1830                   & 0.3288                  & 0.2044                   & 0.2212                  & 0.2227                  \\
                                             & 768 (3-Shot)                                               & 0.2557                  & 0.1381                   & \textbf{0.3225}                  & \textbf{0.2030}                   & 0.2364                  & 0.3073                  \\ \bottomrule
\end{tabular}
}
\caption{We compared two recently proposed prompt methods (LLMTime and LLMP) for zero-shot forecasting and two additional prompts (TS-CoT and TS-InContext) using LLM prompting techniques (Chain-of-Thought and In-Context learning), specifically designed for time-series forecasting. LLMTime demonstrates superior performance compared to other prompts with minimal input. In this experiment, we use GPT-4o as a LLM backbone.}
\label{apptab:comp-prompts}
\vspace{-5mm}
\end{table}

%% file: Figures/desc_LLMTime_Prompt.tex
\begin{figure}[h!]
    \centering
    \begin{minipage}{0.42\textwidth}
        \fbox{
            \begin{minipage}{\textwidth}
                \fontsize{10}{10}\selectfont
                \textbf{(a) LLMTime for base models} \\
                \rule{\textwidth}{0.4pt}
                \fontsize{7}{8}\selectfont                
                -12, -13, -15, -7, -11, -6, 43, 98, 43, -10, -11, -9, -11, -12, -9, -10, -12, -8, -9, -13, 
                \rule{\textwidth}{0.4pt}
                \fontsize{7}{8}\selectfont
                \textbf{Response}: -9, -10, -12, -8, -9
            \end{minipage}
        }
    \vspace{3mm}
    \end{minipage}
    \begin{minipage}{0.42\textwidth}
        \fbox{
            \begin{minipage}{\textwidth}
                \fontsize{10}{10}\selectfont
                \textbf{(b) LLMTime for instruction-tuned models} \\
                \rule{\textwidth}{0.4pt}
                \fontsize{8}{8}\selectfont
                \textbf{System Prompt} \\
                \fontsize{7}{8}\selectfont
                You are a helpful assistant that performs time series predictions. The user will provide a sequence and you will predict the remaining sequence. The sequence is represented by decimal strings separated by commas. \\
                \\
                \fontsize{8}{8}\selectfont
                \textbf{User Prompt} \\
                \fontsize{7}{8}\selectfont
                Please predict next sequence following input sequence without producing any additional text. Do not say anything like 'the next terms in the sequence are', just return the numbers. Input Sequence: -12, -13, -15, -7, -11, -6, 43, 98, 43, -10, -11, -9, -11, -12, -9, -10, -12, -8, -9, -13, 
               \rule{\textwidth}{0.4pt}
                \fontsize{7}{8}\selectfont
                \textbf{Response}: -9, -10, -12, -8, -9
            \end{minipage}
        }
    \end{minipage}
    
    \caption{Two prompts proposed by LLMTime for (a) base LLM models (LLaMA2-7B, 70B, and LLaMA3.1-70B) and (b) instruction-tuned LLM models (GPT-3.5, GPT-4, GPT-4o, and LLaMA-Intstruct).}
    \label{app:llmtime-prompt}
\end{figure}

%% file: Figures/desc_LLMP_Prompt.tex
\begin{figure}[h!]
    \centering
    \begin{minipage}{0.42\textwidth}
        \fbox{
            \begin{minipage}{\textwidth}
                \fontsize{10}{10}\selectfont
                \textbf{(a) LLMP-Independent~(Multi-turn)} \\
                \rule{\textwidth}{0.4pt}
                \fontsize{8}{8}\selectfont
                \textbf{System Prompt} \\
                \fontsize{7}{8}\selectfont
                \\
                \fontsize{8}{8}\selectfont
                \textbf{User Prompt} \\
                \fontsize{7}{8}\selectfont
                0,-12\textbackslash n 1,-13\textbackslash n 2,-15\textbackslash n 3,-7\textbackslash n 4,-11\textbackslash n 5,-6\textbackslash n 6,43\textbackslash n 7,98\textbackslash n 8,43\textbackslash n 9,-10\textbackslash n 10,-11\textbackslash n 11,-9\textbackslash n 12,-11\textbackslash n 13,-12\textbackslash n 14,-9\textbackslash n 15,-10\textbackslash n 16,-12\textbackslash n 17,-8\textbackslash n 18,-9\textbackslash n 19,-13\textbackslash n 20, 
                \rule{\textwidth}{0.4pt}
                \fontsize{7}{8}\selectfont
                \textbf{Response}: -9\\
\rule{\textwidth}{0.4pt}
                \fontsize{8}{8}\selectfont
                \textbf{User Prompt} \\
                \fontsize{7}{8}\selectfont
                0,-12\textbackslash n 1,-13\textbackslash n 2,-15\textbackslash n 3,-7\textbackslash n 4,-11\textbackslash n 5,-6\textbackslash n 6,43\textbackslash n 7,98\textbackslash n 8,43\textbackslash n 9,-10\textbackslash n 10,-11\textbackslash n 11,-9\textbackslash n 12,-11\textbackslash n 13,-12\textbackslash n 14,-9\textbackslash n 15,-10\textbackslash n 16,-12\textbackslash n 17,-8\textbackslash n 18,-9\textbackslash n 19,-13\textbackslash n 21, 
                \rule{\textwidth}{0.4pt}
                \fontsize{7}{8}\selectfont
                \textbf{Response}: -10\\
\rule{\textwidth}{0.4pt}
                \fontsize{8}{8}\selectfont
                \textbf{User Prompt} \\
                \fontsize{7}{8}\selectfont
                0,-12\textbackslash n 1,-13\textbackslash n 2,-15\textbackslash n 3,-7\textbackslash n 4,-11\textbackslash n 5,-6\textbackslash n 6,43\textbackslash n 7,98\textbackslash n 8,43\textbackslash n 9,-10\textbackslash n 10,-11\textbackslash n 11,-9\textbackslash n 12,-11\textbackslash n 13,-12\textbackslash n 14,-9\textbackslash n 15,-10\textbackslash n 16,-12\textbackslash n 17,-8\textbackslash n 18,-9\textbackslash n 19,-13\textbackslash n 22, 
                \rule{\textwidth}{0.4pt}
                \fontsize{7}{8}\selectfont
                \textbf{Response}: -12\\
\rule{\textwidth}{0.4pt}
                \fontsize{8}{8}\selectfont
                \textbf{User Prompt} \\
                \fontsize{7}{8}\selectfont
                0,-12\textbackslash n 1,-13\textbackslash n 2,-15\textbackslash n 3,-7\textbackslash n 4,-11\textbackslash n 5,-6\textbackslash n 6,43\textbackslash n 7,98\textbackslash n 8,43\textbackslash n 9,-10\textbackslash n 10,-11\textbackslash n 11,-9\textbackslash n 12,-11\textbackslash n 13,-12\textbackslash n 14,-9\textbackslash n 15,-10\textbackslash n 16,-12\textbackslash n 17,-8\textbackslash n 18,-9\textbackslash n 19,-13\textbackslash n 23, 
                \rule{\textwidth}{0.4pt}
                \fontsize{7}{8}\selectfont
                \textbf{Response}: -8 \\
\rule{\textwidth}{0.4pt}
                \fontsize{8}{8}\selectfont
                \textbf{User Prompt} \\
                \fontsize{7}{8}\selectfont
                0,-12\textbackslash n 1,-13\textbackslash n 2,-15\textbackslash n 3,-7\textbackslash n 4,-11\textbackslash n 5,-6\textbackslash n 6,43\textbackslash n 7,98\textbackslash n 8,43\textbackslash n 9,-10\textbackslash n 10,-11\textbackslash n 11,-9\textbackslash n 12,-11\textbackslash n 13,-12\textbackslash n 14,-9\textbackslash n 15,-10\textbackslash n 16,-12\textbackslash n 17,-8\textbackslash n 18,-9\textbackslash n 19,-13\textbackslash n 24, 
                \rule{\textwidth}{0.4pt}
                \fontsize{7}{8}\selectfont
                \textbf{Response}: -9
            \end{minipage}
        }
    \vspace{3mm}
    \end{minipage}
    \begin{minipage}{0.42\textwidth}
        \fbox{
            \begin{minipage}{\textwidth}
                \fontsize{10}{10}\selectfont
                \textbf{(b) LLMP-Independent~(Single-turn)} \\
                \rule{\textwidth}{0.4pt}
                \fontsize{8}{8}\selectfont
                \textbf{System Prompt} \\
                \fontsize{7}{8}\selectfont
                 You are a helpful assistant that performs time series predictions. The user will provide you with a sequence of ordered pairs (x, y), and you will predict y for pairs where only x is given. Each pair is separated by a newline. \\
                \\
                \fontsize{8}{8}\selectfont
                \textbf{User Prompt} \\
                \fontsize{7}{8}\selectfont
                Please predict the missing values in the y column based on the given x and y data points without producing any additional text. Do not say anything like 'the next terms in the sequence are', just return only the y values  as numbers without x values. x, y\textbackslash n 0,-12\textbackslash n 1,-13\textbackslash n 2,-15\textbackslash n 3,-7\textbackslash n 4,-11\textbackslash n 5,-6\textbackslash n 6,43\textbackslash n 7,98\textbackslash n 8,43\textbackslash n 9,-10\textbackslash n 10,-11\textbackslash n 11,-9\textbackslash n 12,-11\textbackslash n 13,-12\textbackslash n 14,-9\textbackslash n 15,-10\textbackslash n 16,-12\textbackslash n 17,-8\textbackslash n 18,-9\textbackslash n 19,-13\textbackslash n 20, \textbackslash n 21, \textbackslash n 22, \textbackslash n 23, \textbackslash n 24, \textbackslash n 
               \rule{\textwidth}{0.4pt}
                \fontsize{7}{8}\selectfont
                \textbf{Response}: -9, -10, -12, -8, -9
            \end{minipage}
        }
    \end{minipage}
    
    \caption{(a) is the original prompt proposed in LLMP. The (a) prompt requires independently processing $O$ queries to obtain $O$ output values. This significantly increases the computational cost of using LLMs. To address this issue, we designed a single-turn prompt (b) that enables LLMP to be performed more efficiently.}
    \label{app:LLMP-prompt}
\end{figure}

%% file: Figures/desc_CoT_InContext.tex
\begin{figure}[h!]
    \centering
    \begin{minipage}{0.42\textwidth}
        \fbox{
            \begin{minipage}{\textwidth}
                \fontsize{10}{10}\selectfont
                \textbf{(a) TS-CoT} \\
                \rule{\textwidth}{0.4pt}
                \fontsize{8}{8}\selectfont
                \textbf{System Prompt} \\
                \fontsize{7}{8}\selectfont
                You are a helpful assistant that performs time series predictions. The user will provide a sequence and you will predict the remaining sequence. The sequence is represented by decimal strings separated by commas. \\
                \\
                \fontsize{8}{8}\selectfont
                \textbf{User Prompt} \\
                \fontsize{7}{8}\selectfont
                Sequence during the input period:
                -12, -13, -15, -7, -11, -6, 43, 98, 43, -10, -11, -9, -11, -12, -9, -10, -12, -8, -9, -13, \\
                Let's think step by step.\\\\
                \textbf{Step 1)} Describe the solution process to make future predictions that reflect the description in up to five sentences.\\\\
                \textbf{Step 2)} Considering the answers to previous steps, please predict next sequence following input sequence without producing any additional text. Do not say anything like 'the next terms in the sequence are', just return the numbers. Input Sequence:-12, -13, -15, -7, -11, -6, 43, 98, 43, -10, -11, -9, -11, -12, -9, -10, -12, -8, -9, -13, 
               \rule{\textwidth}{0.4pt}
                \fontsize{7}{8}\selectfont
                \textbf{Response}:\\
                \textbf{Answer 1)}  The sequence provided appears to be non-linear and may contain multiple patterns or regimes. To predict future data points, it's important to identify any apparent periodicity, seasonality, trends, or cycles within the data. The sequence might show different behaviors at different segments, which can include rising trends, falling trends, and various plateaus or periods of relative stability ... \\
                \textbf{Answer 2)} -9, -10, -12, -8, -9
            \end{minipage}
        }
    \vspace{3mm}
    \end{minipage}
    \begin{minipage}{0.42\textwidth}
        \fbox{
            \begin{minipage}{\textwidth}
                \fontsize{10}{10}\selectfont
                \textbf{(b) TS-InContext (3-Shot)} \\
                \rule{\textwidth}{0.4pt}
                \fontsize{8}{8}\selectfont
                \textbf{System Prompt} \\
                \fontsize{7}{8}\selectfont
                You are a helpful assistant that performs time series predictions. The user will provide a sequence and you will predict the remaining sequence. The sequence is represented by decimal strings separated by commas. \\
                \\
                \fontsize{8}{8}\selectfont
                \textbf{User Prompt} \\
                \fontsize{7}{8}\selectfont
                We give you input and output sequence samples:\\
                1. Sequence: -12, -13, -15, -7, -11, <sep>  -6, 43, 98, 43, -10\\
                2. Sequence: -6, 43, 98, 43, -10, <sep> -11, -9, -11, -12, -9\\
                3. Sequence:  -11, -9, -11, -12, -9, <sep> -10, -12, -8, -9, -13\\
                
                Please predict next sequence following input sequence without producing any additional text. Do not say anything like 'the next terms in the sequence are', just return the numbers. Input Sequence: -10, -12, -8, -9, -13, <sep>
               \rule{\textwidth}{0.4pt}
                \fontsize{7}{8}\selectfont
                \textbf{Response}: -9, -10, -12, -8, -9
            \end{minipage}
        }
    \end{minipage}
    
    \caption{(a) TS-CoT uses a step-by-step reasoning approach (CoT) to guide predictions by analyzing patterns before forecasting. (b) TS-InContext (3-Shot) leverages in-context learning by providing examples of input-output pairs, allowing the model to infer patterns directly from demonstrations.}
    \label{app:cot-incontext}
\end{figure}

%% file: Tables/exp_temperature.tex
\begin{table}[ht]
\resizebox{\columnwidth}{!}{
\centering
\begin{tabular}{l|c|r r| r r| r r}
\toprule
\multicolumn{1}{c}{\multirow{2}{*}{\textbf{Models}}} & \multicolumn{1}{c}{\multirow{2}{*}{\textbf{Temp. }}} & \multicolumn{2}{c|}{\textbf{ETTm2}}                & \multicolumn{2}{c|}{\textbf{Exchange Rate}}        & \multicolumn{2}{c}{\textbf{Weather}}              \\ \cmidrule(l){3-8} 
\multicolumn{1}{c}{}                        & \multicolumn{1}{c}{}                              & \multicolumn{1}{c}{MAE} & \multicolumn{1}{c|}{MSE} & \multicolumn{1}{c}{MAE} & \multicolumn{1}{c|}{MSE} & \multicolumn{1}{c}{MAE} & \multicolumn{1}{c}{MSE} \\ \midrule
GPT-4o          & 0.1         & 0.5255      & 0.9171      & 0.4823         & 0.4568         & 0.2356        & 0.1662        \\
GPT-4o          & 0.5         & 0.3057      & 0.2383      & 0.3487         & 0.2631         & 0.2463        & 0.1988        \\
GPT-4o  & 1.0         & 0.2322      & 0.1237      & 0.3595         & 0.2893         & 0.1880        & 0.1147        \\  \midrule
DLinear-S       & –           & 0.2039      & 0.0782      & \textbf{0.2296}         & \textbf{0.1166}         & 0.1643        & 0.0522        \\
RLinear-S       & –           & \textbf{0.1824}      & \textbf{0.0677}      & 0.2571         & 0.1370         & \textbf{0.1392}        & \textbf{0.0469}        \\
\bottomrule
\end{tabular}
}
\caption{Forecasting performance by temperatures}
\label{tab:temp}
\end{table}

%% file: Tables/exp_topp.tex
\begin{table}[ht]
\resizebox{\columnwidth}{!}{
\centering
\begin{tabular}{l|c|rr|rr|rr}
\toprule
\multicolumn{1}{c}{\multirow{2}{*}{\textbf{Models}}} & \multicolumn{1}{c}{\multirow{2}{*}{\textbf{Top-p}}} & \multicolumn{2}{c|}{\textbf{ETTm2}}                & \multicolumn{2}{c|}{\textbf{Exchange Rate}}        & \multicolumn{2}{c}{\textbf{Weather}}              \\ 
\cmidrule(l){3-8} 
\multicolumn{1}{c}{}                        & \multicolumn{1}{c}{}                              & \multicolumn{1}{c}{MAE} & \multicolumn{1}{c|}{MSE} & \multicolumn{1}{c}{MAE} & \multicolumn{1}{c|}{MSE} & \multicolumn{1}{c}{MAE} & \multicolumn{1}{c}{MSE} \\ 
\midrule
GPT-4o          & 0.1         & 0.2816      & 0.2032      & 0.3697         & 0.2930         & 0.1902        & 0.1170        \\
GPT-4o          & 0.5         & 0.3153      & 0.2424      & 0.3913         & 0.3078         & 0.2019        & 0.1235        \\
GPT-4o          & 0.8         & 0.2322      & 0.1237      & 0.3595         & 0.2893         & 0.1880        & 0.1147        \\
GPT-4o          & 1.0         & 0.2825      & 0.1996      & 0.3562         & 0.2926         & 0.1897        & 0.1268        \\  
\midrule
DLinear-S       & –           & 0.2039      & 0.0782      & \textbf{0.2296}         & \textbf{0.1166}         & 0.1643        & 0.0522        \\
RLinear-S       & –           & \textbf{0.1824}      & \textbf{0.0677}      & 0.2571         & 0.1370         & \textbf{0.1392}        & \textbf{0.0469}        \\
\bottomrule
\end{tabular}
}
\caption{Forecasting performance by top-p values}
\label{tab:topp}
\end{table}

%% file: Tables/exp_entire_horizon.tex
\begin{table}[ht]
\resizebox{\columnwidth}{!}{
\centering
\begin{tabular}{l|rr|rr|rr}
\toprule
\multirow{2}{*}{\textbf{Models}} & \multicolumn{2}{c|}{\textbf{ETTm2}}                & \multicolumn{2}{c|}{\textbf{Exchange Rate}}        & \multicolumn{2}{c}{\textbf{Weather}}              \\ 
\cmidrule(l){2-7}
                                  & \textbf{MAE} & \textbf{MSE} & \textbf{MAE} & \textbf{MSE} & \textbf{MAE} & \textbf{MSE} \\ 
\midrule
GPT-4o                             & 0.3391       & 0.2156       & 0.3292       & 0.1824       & 0.0350       & 0.0024       \\
LLaMA3.1-8B-Inst.               & 0.3684       & 0.2508       & 0.3619       & 0.2682       & 0.0365       & 0.0026       \\ \midrule
DLinear-S                          & 0.1455       & 0.0425       & \textbf{0.2728}       & \textbf{0.1105}       & 0.0242       & \textbf{0.0010}       \\
RLinear-S                          & \textbf{0.1385}       &  \textbf{0.0411}       & 0.2974       & 0.1268       & \textbf{0.0235}       & \textbf{0.0010}       \\
\bottomrule
\end{tabular}
}
\caption{Forecasting performance on three real-world datasets over the entire test set range.}
\label{tab:entire_horizon}
\end{table}

%% file: Tables/exp_freq_noise.tex
\begin{table}[ht]
\resizebox{\columnwidth}{!}{
\centering
\begin{tabular}{l|rr|rr|rr}
\toprule
\multirow{2}{*}{\textbf{Models}} & \multicolumn{2}{c|}{\textbf{Clean}} & \multicolumn{2}{c|}{\textbf{Freq-Replace}} & \multicolumn{2}{c}{\textbf{Freq-Add}} \\ 
\cmidrule(l){2-7}
                                  & \textbf{MAE} & \textbf{MSE} & \textbf{MAE}       & \textbf{MSE}       & \textbf{MAE}     & \textbf{MSE}     \\ 
\midrule
GPT-4o                            & \textbf{0.6726}       & \textbf{0.9089}       & 0.9065             & 1.5183             & 0.7862           & 1.2615           \\
GPT-4                             & 0.6956       & 0.9533       & 0.9171             & 1.4361             & 0.7693           & 1.0061           \\ 
\midrule
DLinear-S                         & 0.7074       & 0.8890       & \textbf{0.7059}             & 0.8899             & \textbf{0.7081}           & 0.8902           \\
RLinear-S                         & 0.7182       & 0.8804       & 0.7129             & \textbf{0.8733}             & 0.7183           & \textbf{0.8743}           \\
\bottomrule
\end{tabular}
}
\caption{Forecasting results of each model on the Monash dataset with two periodic noises.}
\label{tab:monash_freq_noise}
\end{table}

%% file: Tables/exp_more_baselines.tex
\begin{table}[ht]
\resizebox{\columnwidth}{!}{
\centering
\begin{tabular}{l|rr|rr|rr}
\toprule
\multicolumn{1}{c}{\multirow{2}{*}{\textbf{Models}}}  & \multicolumn{2}{c|}{\textbf{ETTm2}}                & \multicolumn{2}{c|}{\textbf{Exchange Rate}}        & \multicolumn{2}{c}{\textbf{Weather}}              \\ 
\cmidrule(l){2-7} 
\multicolumn{1}{c}{}                                                    & \multicolumn{1}{c}{MAE} & \multicolumn{1}{c|}{MSE} & \multicolumn{1}{c}{MAE} & \multicolumn{1}{c|}{MSE} & \multicolumn{1}{c}{MAE} & \multicolumn{1}{c}{MSE} \\ 
\midrule
GPT-4o                 & 0.2322      & 0.1237      & 0.3595         & 0.2893         & 0.1880        & 0.1147        \\ \midrule
ARIMA                    & 0.3402      & 0.2201      & 0.3460         & 0.2535         & 0.3104        & 0.2722        \\
N-BEATS                  & 0.2804      & 0.1842      & 0.3808         & 0.3079         & 0.2169        & 0.1209        \\ \midrule
DLinear-S               & 0.2039      & 0.0782      & \textbf{0.2296}         & \textbf{0.1166}         & 0.1643        & 0.0522        \\
RLinear-S              & \textbf{0.1824}      & \textbf{0.0677}      & 0.2571         & 0.1370         & \textbf{0.1392}        & \textbf{0.0469}        \\
\bottomrule
\end{tabular}
}
\caption{Forecasting results of ARIMA and N-BEATS on three real-world datasets}
\label{tab:more_baselines}
\end{table}

%% file: Figures/appendix_Electricity_vis.tex
\begin{figure*}[h!]
\begin{center}
  \includegraphics[width=1.0\linewidth]{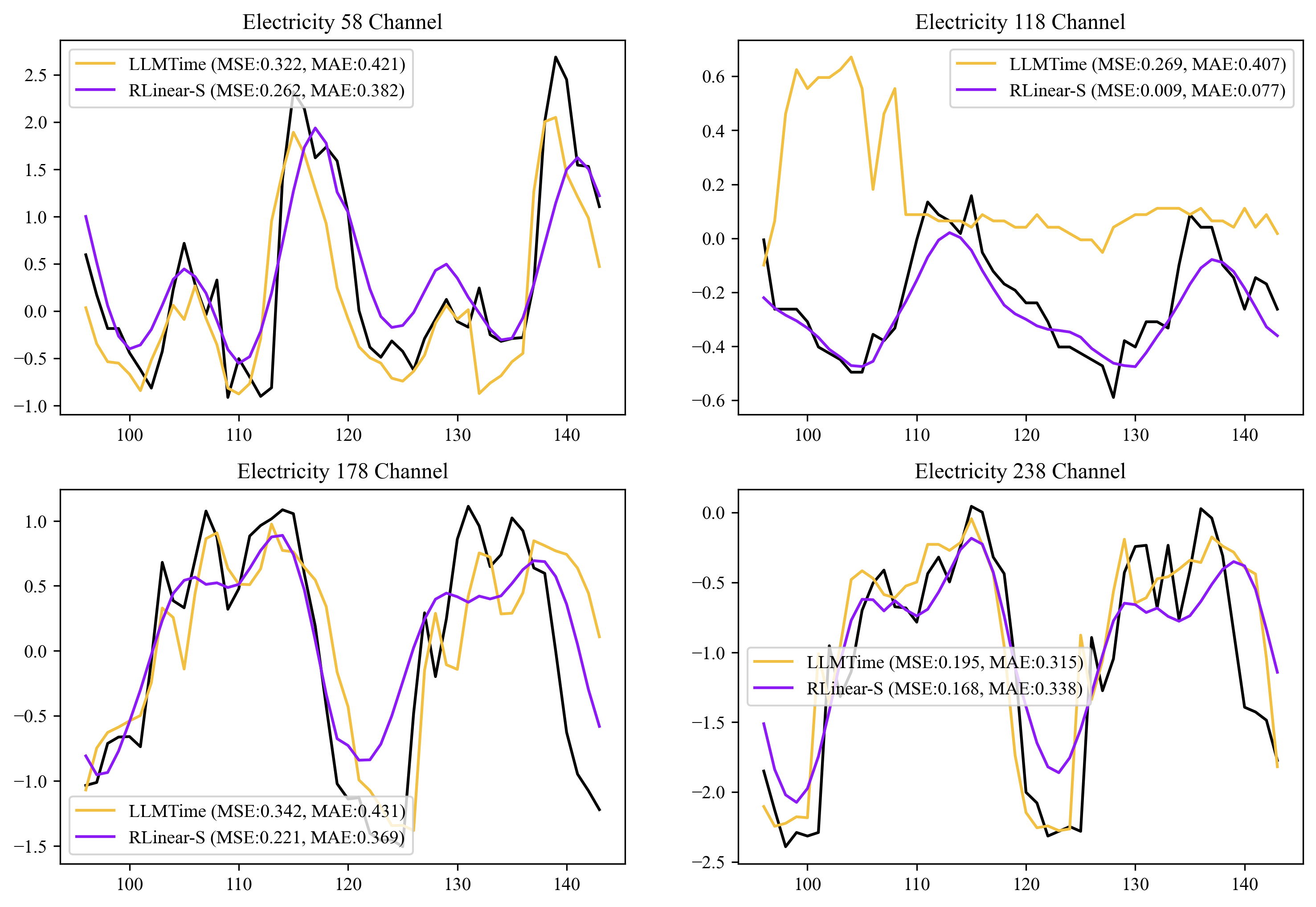}
\end{center}
\vspace{-3mm}
\caption{Qualitative results on Electricity dataset across four channels (58, 118, 178, and 238). The black line represents ground truth, while LLMTime (yellow) and RLinear-S (purple) show model predictions with corresponding MSE and MAE values. Overall, RLinear-S tends to outperform LLMTime.}
\label{fig:app-ecl-vis}
\end{figure*}

%% file: Figures/appendix_ETTm2_vis.tex
\begin{figure*}[h!]
\begin{center}
  \includegraphics[width=1.0\linewidth]{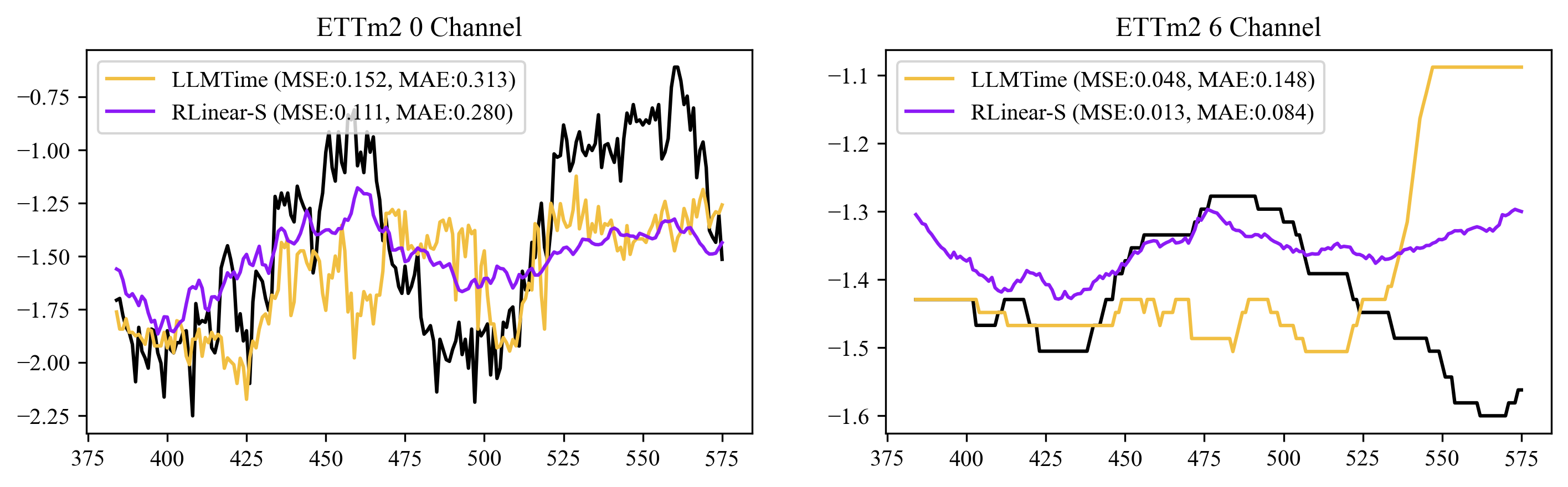}
\end{center}
\vspace{-3mm}
\caption{Qualitative results on ETTm2 dataset across four channels (0 and 6}
\label{fig:app-ettm2-vis}
\end{figure*}

%% file: Figures/appendix_ExchangeRate_vis.tex
\begin{figure*}[h!]
\begin{center}
  \includegraphics[width=1.0\linewidth]{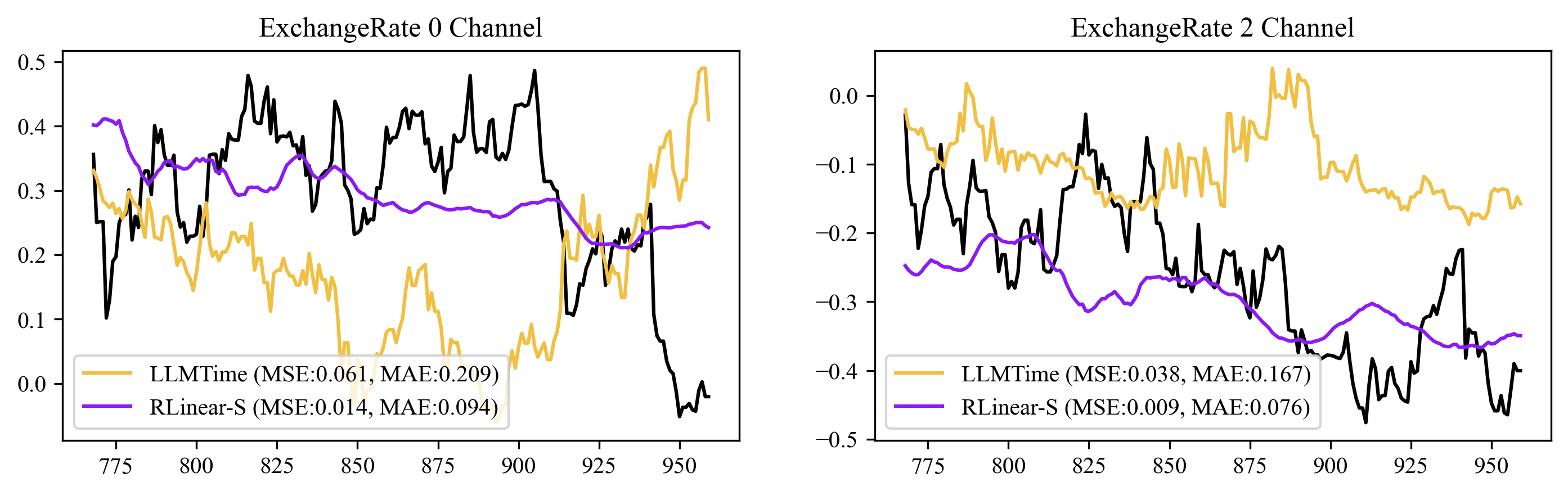}
\end{center}
\vspace{-3mm}
\caption{Qualitative results on ExchangeRate dataset across four channels (0 and 2)}
\label{fig:app-ettm2-vis}
\end{figure*}

%% file: Figures/appendix_Weather_vis.tex
\begin{figure*}[h!]
\begin{center}
  \includegraphics[width=1.0\linewidth]{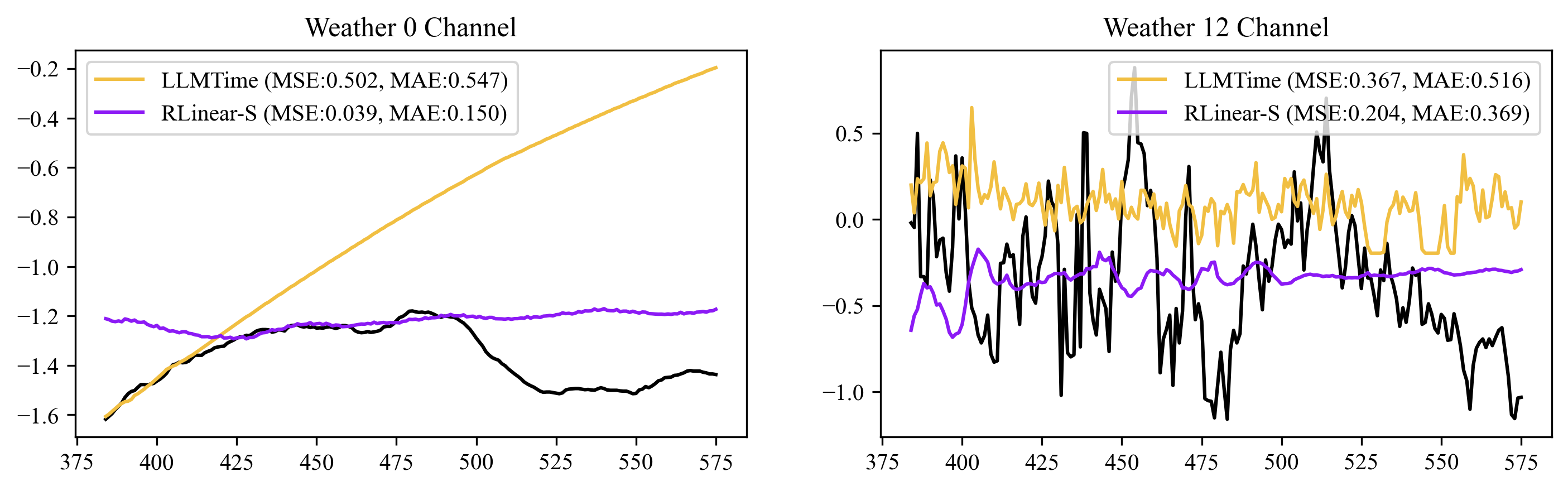}
\end{center}
\vspace{-3mm}
\caption{Qualitative results on Weather dataset across four channels (0 and 12)}
\label{fig:app-weather-vis}
\end{figure*}

%% file: Figures/appendix_Traffic_vis.tex
\begin{figure*}[h!]
\begin{center}
  \includegraphics[width=1.0\linewidth]{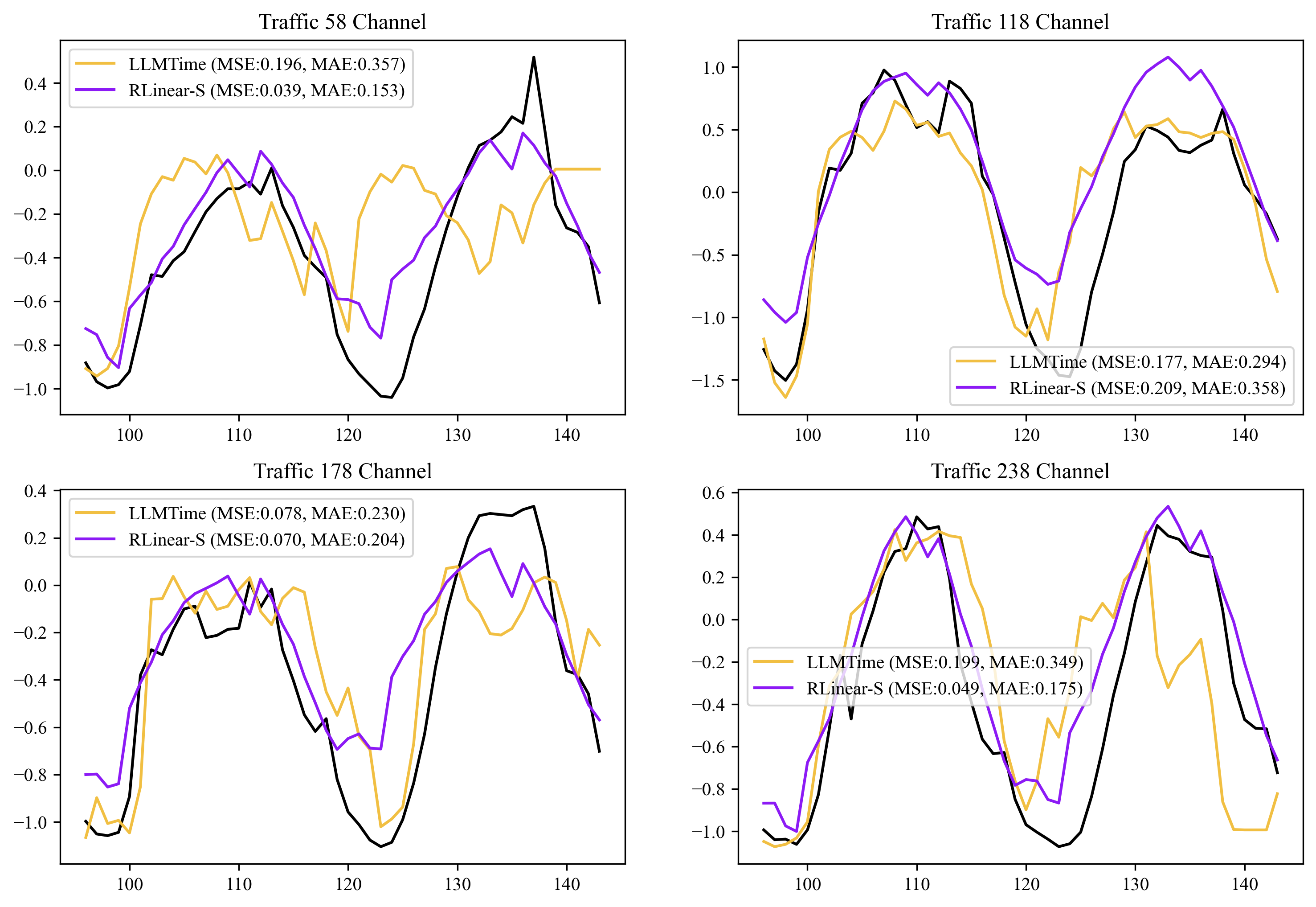}
\end{center}
\vspace{-3mm}
\caption{Qualitative results on Traffic dataset across four channels (58, 118, 178, and 238).}
\label{fig:app-ettm2-vis}
\end{figure*}

%% file: Figures/appendix_full_results.tex
\begin{figure*}[h]
\begin{center}
  \includegraphics[width=0.81\linewidth]{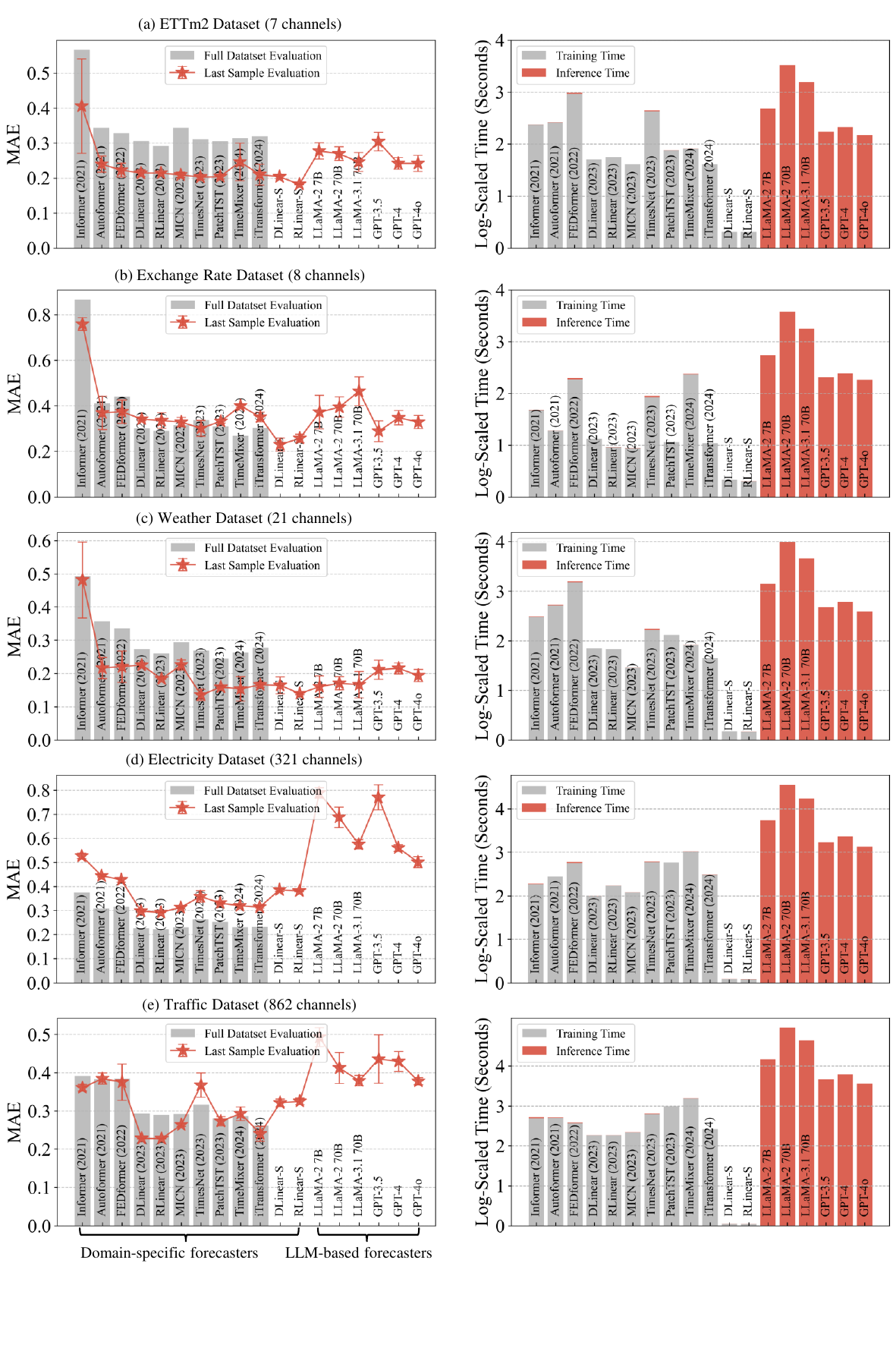}
\end{center}
\vspace{-15mm}
\caption{Multivariate forecasting results on five datasets from the Informer benchmark. We report the normalized MAE and log-scaled inference time for domain-specific and LLM-based forecasters with confidence intervals.}
\label{appfig:full-res}
\end{figure*}